\documentclass{article}

\usepackage{iclr2027_conference,times}

\usepackage{amsmath,amsfonts,bm}

\def\eqref#1{equation~\ref{#1}}

\def\1{\bm{1}}

\DeclareMathAlphabet{\mathsfit}{\encodingdefault}{\sfdefault}{m}{sl}
\SetMathAlphabet{\mathsfit}{bold}{\encodingdefault}{\sfdefault}{bx}{n}

\usepackage{graphicx}
\usepackage{booktabs}
\usepackage{multirow}
\usepackage{hyperref}
\usepackage{url}

\newcommand{\methodname}{\textsc{MoLE}}

\title{\methodname: Mixture of Latent Experts for Complementary Visual Reasoning}

\author{%
{\normalfont
\begin{tabular}[t]{@{}l@{}}
\textbf{Yingcheng Liu$^{1}$ \quad Tianyi Jiang$^{1}$ \quad Yujuan Ding$^{2}$ \quad jiangbo Ai$^{3}$} \\
\textbf{Xun Jiang$^{1}$ \quad Guoqing Wang$^{4}$ \quad Wei Ye$^{1}$ \quad Yi Bin$^{1}$} \\
\\
$^{1}$Tongji University, China \\
$^{2}$Hong Kong Polytechnic University, Hong Kong \\
$^{3}$Alibaba Group, China \\
$^{4}$University of Electronic Science and Technology of China, China
\end{tabular}%
}%
}

\iclrfinalcopy

\begin{document}

\maketitle
\lhead{Preprint}

\begin{abstract}
Latent visual reasoning equips vision--language models with continuous intermediate states that can process visual evidence without explicit textual reasoning traces or repeated image operations. However, existing methods often allow multiple latent tokens to access the same visual evidence through shared value projections, providing no mechanism for them to extract complementary visual information; simply increasing the latent budget can therefore yield redundant latent representations. We argue that effective latent reasoning should encourage different latent tokens to extract complementary visual information, and thereby act as specialized visual experts. Based on this insight, we propose \methodname{}, a Mixture of Latent Experts framework that controls both what visual evidence each latent visual expert observes and how it transforms that evidence. \methodname{} isolates latent visual experts during evidence extraction and uses dedicated latent summary experts to aggregate the complementary representations of latent visual experts. A two-stage training pipeline first forces visual evidence through this latent pathway and then restores direct visual access, requiring neither predefined expert roles nor intermediate visual targets. Across five visual reasoning benchmarks, \methodname{} achieves an average score of 78.6, outperforming data-matched supervised fine-tuning by 4.9 and the strongest evaluated latent visual reasoning baseline at the same latent budget by 3.6. Representation analyses show lower latent-state similarity and more diverse visual attention, while masking the latent pathway reduces average performance by 9.2. These results demonstrate that specializing latent computation is more effective than merely increasing the number of latent tokens.
\end{abstract}

\section{Introduction}
\label{sec:introduction}

Visual reasoning requires a vision--language model (VLM) to generate an answer through intermediate computation over visual inputs. The model must identify visual evidence, preserve fine-grained details, and combine cues that may be distributed across multiple objects, regions, or images. Existing approaches instantiate this intermediate process in different ways. Text-based chain-of-thought expresses it in discrete language \citep{wei2022chain,xu2025llavacot,huang2026visionr1}. Interleaved approaches acquire additional visual observations during reasoning, but often require cropping, re-encoding, or external tools at inference \citep{shao2024visualcot,wang2025pixelreasoner,wu2026vtoolr1}. Latent visual reasoning instead performs the intermediate computation in continuous hidden states \citep{li2026lvr,qin2026covt,li2026livr}. It therefore keeps the reasoning process internal without explicit textual traces or repeated image operations.

However, this advantage does not by itself make multiple latent tokens useful. In a common design, all latents can access the same visual tokens and process them through a shared value projection \citep{li2026lvr,qin2026covt,li2026livr}. These latents may therefore converge to redundant representations. Merely increasing the latent count can add more redundant states rather than new reasoning capacity.

\begin{figure}[t]
  \centering
  \includegraphics[width=\linewidth]{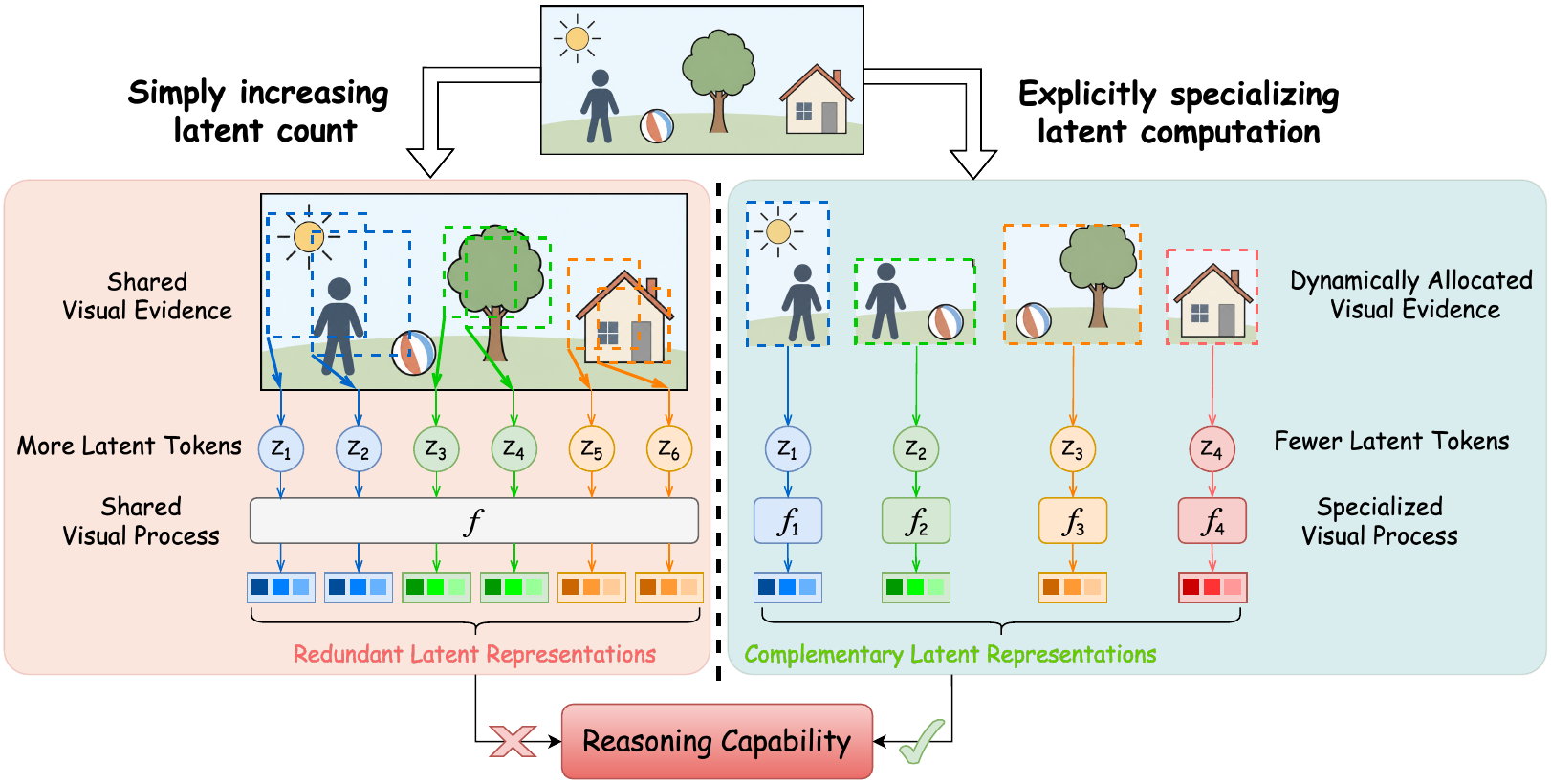}
  \caption{Two paths to enhancing latent visual reasoning. Simply increasing latent count while sharing visual evidence and transformations can produce redundant representations. Explicitly specializing what each latent observes and how it transforms that evidence encourages complementary representations and stronger reasoning without increasing the latent budget.}
  \label{fig:motivation}
\end{figure}

This limitation motivates us to explore explicit specialization as a way to make better use of a fixed latent budget. A latent representation is determined by both \emph{what} visual evidence it receives and \emph{how} it transforms that evidence. When these two factors are shared across all latents, increasing the latent count can largely replicate the same computation and produce redundant representations. Differentiating both the available evidence and its transformation encourages latents to capture complementary visual information within a fixed budget. Figure~\ref{fig:motivation} contrasts this explicit specialization with simply increasing the latent count under shared evidence and transformations.

Motivated by this insight, we introduce \textbf{\methodname{}} (\textbf{M}ixture of \textbf{L}atent \textbf{E}xperts) for complementary visual reasoning. At every decoder layer, sparse visual routing assigns each visual token to a subset of latent visual experts, differentiating the evidence available to them. Each latent visual expert further uses an Expert-Specific Value Adapter (ESVA) to transform its routed visual evidence differently. We isolate latent visual experts from one another to preserve their complementarity, then introduce latent summary experts that aggregate only the representations produced by the latent visual experts. Answer tokens attend to both latent expert sets, allowing complementary visual information and its aggregated summary to jointly support answer prediction.

\methodname{} is trained without predefined expert roles or intermediate visual targets. Because direct visual access can allow answer tokens to bypass the latent states, we follow LIVR \citep{li2026livr} and adopt the two-stage training design to ensure that the latent pathway learns visual information useful for answer generation. Stage~1 imposes a visual attention bottleneck that forces visual evidence through the latent pathway. Stage~2 restores direct visual access, allowing answer generation to combine the learned latent representations with the original visual tokens. Training uses supervision only on answer tokens, together with the MoE loss for router optimization and a diversity objective for the ESVAs. Both stages use this same joint training objective. At inference, the latent states are computed during prefilling and require no additional image processing.

Across five visual reasoning benchmarks, \methodname{} achieves an average score of 78.6, improving data-matched SFT by 4.9 and the strongest evaluated latent visual reasoning baseline at the same latent budget by 3.6.
It produces the lowest latent-state similarity and the highest KL diversity of latent attention-pattern among the compared latent visual reasoning methods. Moreover, masking the learned latent pathway lowers the average by 9.2 despite preserving direct visual access, showing that the answer generation uses information carried by the latent tokens. Doubling the latent budget does not improve the best configuration, further supporting the importance of specializing latent computation rather than merely increasing its size.

Our contributions are as follows:
\begin{itemize}
  \item We identify the lack of complementary latent computation as a central limitation of latent visual reasoning and motivate encouraging latents to extract complementary visual information.
  \item We propose \methodname{}, a new latent visual reasoning method that combines dynamic visual allocation, expert-specific visual transformations, and topological latent aggregation without predefined expert roles or intermediate visual targets.
  \item We demonstrate consistent gains across visual reasoning benchmarks and provide ablation, representation, and intervention evidence that the learned latents are complementary and functionally used for answer generation.
\end{itemize}
\section{Related Work}
\label{sec:related_work}

\noindent\textbf{Text-based visual reasoning.}
Chain-of-thought (CoT) prompting improves reasoning in large language models (LLMs) by eliciting textual intermediate steps before an answer \citep{wei2022chain}. This paradigm extends to vision--language models (VLMs): LLaVA-CoT organizes generation into summarization, visual interpretation, logical reasoning, and conclusion stages \citep{xu2025llavacot}, whereas Vision-R1 combines a multimodal CoT cold start with reinforcement learning \citep{huang2026visionr1}. These methods provide readable task decompositions, but expressing visual reasoning entirely through discrete text may omit fine-grained evidence, motivating methods that operate on visual information more directly.

\noindent\textbf{Interleaved multimodal reasoning.}
Interleaved multimodal reasoning incorporates processed visual observations into textual trajectories. Visual CoT predicts a question-relevant bounding box, then combines re-encoded crop tokens with full-image tokens for answer generation \citep{shao2024visualcot}. Tool-based methods acquire evidence more flexibly: Pixel Reasoner invokes zoom-in and frame-selection operations \citep{wang2025pixelreasoner}; VTool-R1 learns to edit charts and tables with Python tools \citep{wu2026vtoolr1}; and VISTA-R1 coordinates visual and symbolic tools over multi-turn interactions \citep{lu2026vistagym}. These methods adapt evidence acquisition to the question but incur additional tool execution and visual processing at inference.

\noindent\textbf{Latent visual reasoning.}
Latent reasoning instead represents intermediate computation as continuous states. COCONUT, CODI, and SeLaR explore this paradigm in LLMs \citep{hao2025coconut,shen2025codi,fu2026selar}, motivating its extension to visual reasoning without explicit image operations at inference. Latent visual reasoning methods differ mainly in how they supervise latent states. LVR reconstructs query-relevant visual tokens, while LANTERN and VaLR align latents with vision-encoder features \citep{li2026lvr,viveiros2026lantern,jeon2026valr}. Mirage distills latent tokens from ground-truth image embeddings, ILVR selectively distills relevant features from helper images, and Monet aligns observation-token representations under controlled access to auxiliary images \citep{yang2026mirage,dong2026ilvr,wang2026monet}. UniVLR compresses a shared canvas containing rendered reasoning traces and auxiliary images, whereas CoVT supervises distinct token groups with segmentation, depth, edge, and DINO targets \citep{jiang2026univlr,qin2026covt}. In contrast, LIVR trains learnable latent tokens without intermediate targets, using an attention bottleneck and answer-token loss \citep{li2026livr}.

However, these methods generally let multiple latents access the same visual evidence through shared value projections, without explicitly encouraging complementary extraction. Increasing the latent budget can therefore yield redundant representations. \methodname{} instead sparsely routes visual evidence to isolated latent visual experts with expert-specific value transformations, then aggregates their complementary representations through dedicated latent summary experts.
\section{Method}
\label{sec:method}

\subsection{Preliminaries}
\label{sec:preliminaries}

Given a set of images $\mathcal{X}=\{X_1,\ldots,X_M\}$, a question prompt $Q$, and a target answer $Y=(y_1,\ldots,y_T)$, the task is to generate $Y$ conditioned on $\mathcal{X}$ and $Q$. A vision--language model uses a vision encoder to extract image features and a visual projector to map them into the language-model hidden space, yielding $N_v$ visual tokens $\mathbf{V}^{(0)}\in\mathbb{R}^{N_v\times d}$. Separately, a text tokenizer converts $Q$ and $Y$ into token ids, which the language model's input embedding layer then maps to prompt tokens $\mathbf{Q}^{(0)}\in\mathbb{R}^{N_q\times d}$ and answer tokens $\mathbf{Y}^{(0)}\in\mathbb{R}^{T\times d}$, respectively. A latent visual reasoning model then introduces $N_z$ latent tokens between the prompt and answer:
\begin{equation}
\mathbf{Z}
= [\mathbf{z}_1,\ldots,\mathbf{z}_{N_z}]
\in\mathbb{R}^{N_z\times d}.
\label{eq:generic_latents}
\end{equation}
The resulting training sequence concatenates these token embeddings in the following order:
\begin{equation}
\mathbf{H}^{(0)}
= [\mathbf{V}^{(0)};\mathbf{Q}^{(0)};\mathbf{Z};\mathbf{Y}^{(0)}].
\label{eq:training_sequence}
\end{equation}
During the prefix forward pass, the latent tokens gather visual and textual context through the decoder attention layers. The model then generates $Y$ autoregressively while attending to the resulting latent representations. This paradigm provides continuous intermediate computation. However, it does not specify which visual evidence each latent token should observe or how different latent tokens should transform that evidence, which can lead to redundant latent representations.

\subsection{Mixture of Latent Experts} 
\label{sec:mixture_latent_experts}

\begin{figure}[t]
  \centering
  \includegraphics[width=\linewidth]{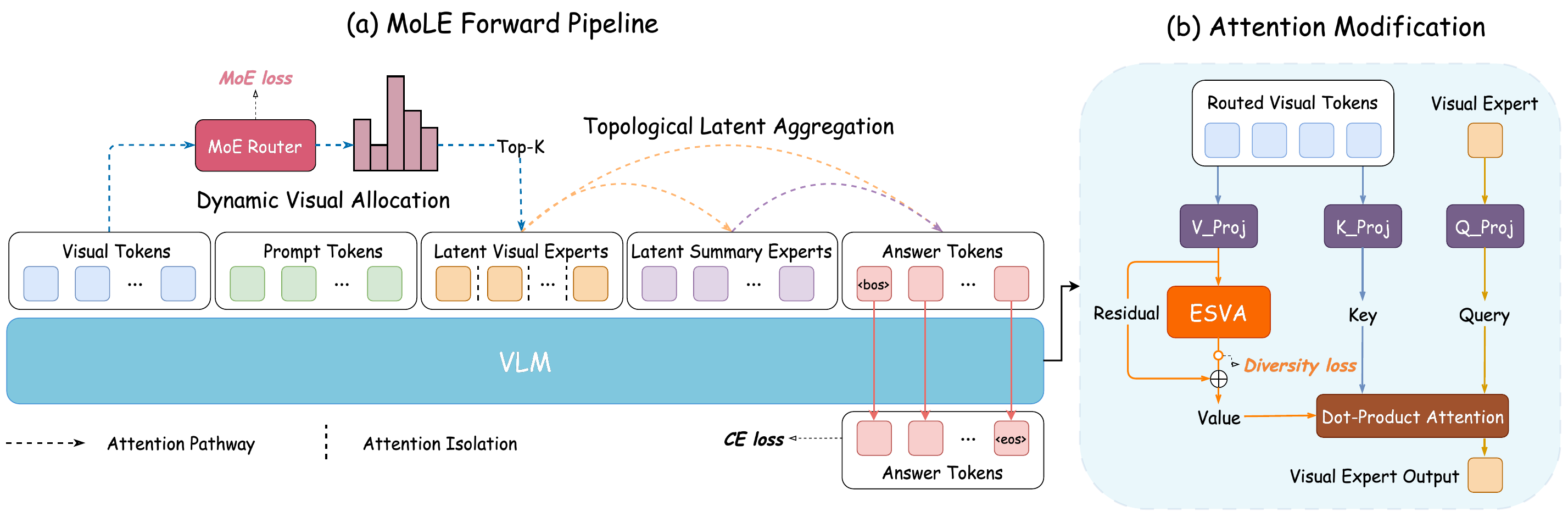}
  \caption{Overview of \methodname{}. (a) A layer-wise router sparsely assigns visual tokens to isolated latent visual experts; latent summary experts aggregate their representations, and answer tokens attend to both latent sets. (b) An Expert-Specific Value Adapter (ESVA) is used within the attention block when a latent visual expert attends to the routed visual tokens. Importantly, the ESVA is activated only for expert--visual-token connections selected by the router; all other attention interactions use the standard attention computation without an ESVA.}
  \label{fig:mole_architecture}
\end{figure}

We introduce the components of \methodname{} by tracing the path of visual information through the model. We first describe how visual evidence is dynamically allocated and transformed, and then explain how the resulting latent representations are aggregated. Subsequently we present the two-stage training pipeline and the training objectives. Figure~\ref{fig:mole_architecture} provides an overview of the framework.

\noindent\textbf{Dynamic visual allocation.}
To distribute visual evidence across latent computations, \methodname{} introduces $E$ latent visual experts:
\begin{equation}
\mathbf{Z}^{\mathrm{vis}}
= [\mathbf{z}^{\mathrm{vis}}_1,\ldots,\mathbf{z}^{\mathrm{vis}}_E]
\in\mathbb{R}^{E\times d}.
\label{eq:latent_visual_experts}
\end{equation}
Unlike a standalone feed-forward expert in conventional MoE models \citep{shazeer2017moe,fedus2022switch}, each latent visual expert acts as a latent query in the decoder sequence. It attends to a routed subset of visual tokens and applies an expert-specific transformation to the corresponding visual evidence. To determine this subset, each decoder layer $\ell\in\{1,\ldots,L\}$ uses an independent router $\mathbf{W}^{R}_{\ell}\in\mathbb{R}^{E\times d}$. Let $\mathbf{h}^{(\ell-1)}_v$ denote the hidden state of visual token $v$ entering layer $\ell$. The router computes expert logits and a distribution over all latent visual experts:
\begin{equation}
r_{\ell,v,e}
= (\mathbf{w}^{R}_{\ell,e})^\top\mathbf{h}^{(\ell-1)}_v,
\qquad
\pi_{\ell,v,e}
= \frac{\exp(r_{\ell,v,e})}{\sum_{j=1}^{E}\exp(r_{\ell,v,j})}.
\label{eq:router_logits}
\end{equation}
Each visual token selects its $K$ highest-scoring latent visual experts, $\mathcal{S}_{\ell,v}=\operatorname{TopK}(\{r_{\ell,v,e}\}_{e=1}^{E},K)$.
If token $v$ is assigned to latent visual expert $e$, the latent visual expert's latent query can attend to that token. All unselected expert--token edges are masked. Each latent visual expert consequently receives a dynamically allocated subset of the visual evidence at each decoder layer.

\noindent\textbf{Router prior steering.}
Top-$K$ selection determines whether a latent visual expert can access a visual token. Within the selected expert set for a visual token, we first renormalize the router logits:
\begin{equation}
\rho_{\ell,v,e}
= \frac{\exp(r_{\ell,v,e})}
{\sum_{j\in\mathcal{S}_{\ell,v}}\exp(r_{\ell,v,j})},
\qquad e\in\mathcal{S}_{\ell,v},
\label{eq:selected_router_prob}
\end{equation}
We then convert the normalized score into an additive prior, $b_{\ell,v,e}=\alpha\log(K\rho_{\ell,v,e}+\varepsilon)$, where $\varepsilon>0$ is a small constant that prevents taking the logarithm of zero and ensures numerical stability. We add this prior to the attention logit when the corresponding latent visual expert attends to that visual token. For attention head $h$,
\begin{equation}
\widetilde a^{(\ell,h)}_{e,v}
=\dfrac{(\mathbf{q}^{(\ell,h)}_{z^{\mathrm{vis}}_e})^\top
\mathbf{k}^{(\ell,h)}_v}{\sqrt{d_h}}+b_{\ell,v,e},
\qquad e\in\mathcal{S}_{\ell,v}.
\label{eq:routed_attention}
\end{equation}
Here, $\mathbf{q}^{(\ell,h)}_{z^{\mathrm{vis}}_e}$ is the query vector of latent visual expert $e$, and $\mathbf{k}^{(\ell,h)}_v$ is the key vector of visual token $v$, both at decoder layer $\ell$ and attention head $h$. The head dimension is denoted by $d_h$. The modified logits are then normalized together with the other accessible attention logits through softmax, producing the attention weights used to aggregate the corresponding value representations. The factor $K$ centers the prior near zero when routing is uniform within the selected expert set. The router can therefore learn which visual experts are better suited to which visual tokens.

\noindent\textbf{Expert-specific value adapter.}
Routing differentiates which visual evidence is available to each latent visual expert. Without expert-specific processing, however, all experts would still apply the same value projection and may produce redundant latent representations. We therefore equip each latent visual expert with an Expert-Specific Value Adapter (ESVA), implemented as an expert-specific two-layer MLP. For visual token $v$ at layer $\ell$, we write the transformation compactly as
\begin{equation}
\mathbf{u}_{\ell,v}
=\operatorname{flatten}(\mathbf{W}^{V}_{\ell}\mathbf{h}^{(\ell-1)}_v),
\qquad
\boldsymbol\delta_{\ell,v,e}
=\operatorname{ESVA}_e(\mathbf{u}_{\ell,v}),
\qquad
\widetilde{\mathbf{u}}_{\ell,v,e}
=\boldsymbol\delta_{\ell,v,e}+\gamma_{\ell,e}\mathbf{u}_{\ell,v}.
\label{eq:esva}
\end{equation}
Here, $\mathbf{W}^{V}_{\ell}$ is the value-projection matrix in the attention block of decoder layer $\ell$, and $\mathbf{h}^{(\ell-1)}_v$ is the hidden state of visual token $v$ entering that layer. The projected value is flattened across attention heads to form $\mathbf{u}_{\ell,v}$, while $\gamma_{\ell,e}$ is a learned residual scale. The resulting $\widetilde{\mathbf{u}}_{\ell,v,e}$ serves as the new value representation for the routed expert--token connection and is used in the subsequent attention-weighted value aggregation. ESVAs bring different visual information transformations to various latent visual experts. The detailed ESVA architecture is provided in Appendix~\ref{app:esva_details}.

\noindent\textbf{Topological latent aggregation.}
After the latent visual experts extract complementary visual information, \methodname{} introduces $S$ latent summary experts to aggregate their representations:
\begin{equation}
\mathbf{Z}^{\mathrm{sum}}
= [\mathbf{z}^{\mathrm{sum}}_1,\ldots,\mathbf{z}^{\mathrm{sum}}_S]
\in\mathbb{R}^{S\times d}.
\label{eq:latent_summary_experts}
\end{equation}
The complete decoder input sequence is therefore $[\mathbf{V}^{(0)};\mathbf{Q}^{(0)};\mathbf{Z}^{\mathrm{vis}};\mathbf{Z}^{\mathrm{sum}};\mathbf{Y}^{(0)}]$. Direct interaction among latent visual experts during evidence extraction could prematurely mix their representations and weaken their complementarity. We therefore block attention between latent visual experts and use the latent summary experts to aggregate their complementary representations into a more complete visual representation. Answer tokens can attend to both visual and summary experts, allowing answer generation to use both the complementary evidence and its aggregated representation.

\noindent\textbf{Two-stage training pipeline.}
Under conventional end-to-end training, answer tokens can directly attend to the original visual tokens, creating a shortcut that allows the model to minimize the answer loss while ignoring the latent pathway. The resulting latent tokens may therefore carry little visual information that is useful for answer generation. Following the two-stage training strategy of LIVR \citep{li2026livr}, we first make the latent pathway informative and then integrate it with direct visual access. In Stage~1, we block direct attention from ordinary prompt and answer tokens to visual tokens, so only latent visual experts can access routed visual evidence. This visual attention bottleneck trains the latent visual experts to extract and preserve task-relevant information. Stage~2 restores direct visual access for prompt and answer tokens, allowing answer generation to rely jointly on the learned latent representations and the original visual tokens. We provide the detailed attention-bottleneck configurations for both stages in Appendix~\ref{app:attention_bottleneck_details}.

\noindent\textbf{Training objectives.}
We first apply language-modeling supervision only to answer tokens:
\begin{equation}
\mathcal{L}_{\mathrm{CE}}
=-\frac{1}{T}\sum_{t=1}^{T}
\log p_{\Theta}
(y_t\mid\mathcal{X},Q,\mathbf{Z}^{\mathrm{vis}},
\mathbf{Z}^{\mathrm{sum}},y_{<t}).
\label{eq:answer_ce}
\end{equation}
Furthermore, we regularize the router with the load-balancing loss used in conventional MoE models \citep{shazeer2017moe,fedus2022switch} and the router $z$-loss introduced by ST-MoE \citep{zoph2022stmoe}. To distinguish the experts selected by Top-$K$ routing from their soft routing probabilities, we define the binary selection variable
\begin{equation}
g_{\ell,v,e}
=
\begin{cases}
1, & \text{if expert } e \text{ is selected for visual token } v
     \text{ at layer } \ell, \\
0, & \text{otherwise}.
\end{cases}
\label{eq:hard_router_selection}
\end{equation}
The selection frequency and mean routing probability of expert $e$ are
\begin{equation}
f_e=\frac{1}{KLN_v}\sum_{\ell=1}^{L}\sum_{v=1}^{N_v}g_{\ell,v,e},
\qquad
\overline{\pi}_e
=\frac{1}{LN_v}\sum_{\ell=1}^{L}\sum_{v=1}^{N_v}\pi_{\ell,v,e}.
\label{eq:router_statistics}
\end{equation}
We define
\begin{equation}
\mathcal{L}_{\mathrm{balance}}
=E\sum_{e=1}^{E}\operatorname{sg}(f_e)\overline{\pi}_e,
\qquad
\mathcal{L}_{z}
=\frac{1}{LN_v}\sum_{\ell=1}^{L}\sum_{v=1}^{N_v}
\left(\log\sum_{e=1}^{E}\exp(r_{\ell,v,e})\right)^2,
\label{eq:router_losses}
\end{equation}
where $\operatorname{sg}$ stops gradients through the discrete selection frequency. The load-balancing loss prevents visual experts from being persistently bypassed and becoming inactive, while the $z$-loss controls the scale of the router logits, thereby stabilizing gradients during training. We refer to their weighted combination, $\mathcal{L}_{\mathrm{MoE}}=\lambda_{\mathrm{balance}}\mathcal{L}_{\mathrm{balance}}+\lambda_z\mathcal{L}_{z}$, as the \emph{MoE loss}.

In addition, to discourage different ESVAs from producing redundant representations for the same visual token, we minimize their pairwise squared cosine similarity. With
$\overline{\boldsymbol\delta}_{\ell,v,e}
=\boldsymbol\delta_{\ell,v,e}/
\lVert\boldsymbol\delta_{\ell,v,e}\rVert_2$, the diversity loss is
\begin{equation}
\mathcal{L}_{\mathrm{div}}
=\frac{1}{LN_v}\sum_{\ell=1}^{L}\sum_{v=1}^{N_v}
\frac{2}{K(K-1)}
\sum_{\substack{i<j\\i,j\in\mathcal{S}_{\ell,v}}}
\left(
\overline{\boldsymbol\delta}_{\ell,v,i}^{\top}
\overline{\boldsymbol\delta}_{\ell,v,j}
\right)^2.
\label{eq:diversity_loss}
\end{equation}
For $K=1$, no selected expert pair exists and we define $\mathcal{L}_{\mathrm{div}}=0$. The complete objective is $\mathcal{L}=\mathcal{L}_{\mathrm{CE}}+\mathcal{L}_{\mathrm{MoE}}+\lambda_{\mathrm{div}}\mathcal{L}_{\mathrm{div}}$. Both training stages use the same joint objective.

\section{Experiments}
\label{sec:experiments}

We adopt Qwen2.5-VL-7B \citep{bai2025qwen25vl} as the backbone VLM. In both training stages, we fine-tune the language backbone with LoRA \citep{hu2022lora} on its attention and MLP blocks, while keeping the vision encoder and projector frozen. In addition to the LoRA parameters, we unfreeze only the embedding rows corresponding to the $E+S$ newly introduced latent experts.

We construct a 63K-example training set by sampling from CoVT-Dataset \citep{qin2026covt}, which combines vision-centric data from LLaVA-OneVision \citep{li2025llavaonevision} with counting and relative-depth samples derived from TallyQA \citep{acharya2019tallyqa} and ADE20K \citep{zhou2017ade20k}. We do not sample directly from any of the evaluation benchmarks described below, thereby avoiding direct reuse of their evaluation splits. Detailed data sampling, processing, and decontamination against the evaluation benchmarks are described in Appendix~\ref{app:dataset_construction}.

\subsection{Evaluation Benchmarks}

We evaluate across five benchmarks with distinct perceptual demands. CVBench \citep{tong2024cambrian} separates 2D spatial relations and object counting from 3D depth ordering and relative-distance understanding. BLINK \citep{fu2024blink} broadens this coverage by recasting 14 classical computer-vision tasks as single- and multi-image multiple-choice questions, including visual correspondence, relative depth, and multi-view reasoning. RealWorldQA (RWQA) \citep{xai2024realworldqa} shifts the evaluation to natural physical scenes, where questions involve properties such as relative size, direction, and navigability. V* \citep{wu2024vstar} focuses specifically on visual search, requiring models to locate and reason about details in high-resolution, visually crowded images. Finally, POPE \citep{li2023pope} complements these capability-oriented benchmarks by probing resistance to object hallucination through binary questions about object presence. For CVBench, we report accuracy separately on its 2D and 3D subsets. We report F1 for POPE and accuracy for the remaining benchmarks. Avg. is the unweighted mean of these six metrics.

\subsection{Main Results}
\label{sec:overall_results}

\noindent\textbf{Baselines.}
We compare \methodname{} with several categories of reasoning models. GPT-4o and Claude-4-Sonnet serve as proprietary VLM references. We also report the Qwen2.5-VL-7B base model and Qwen2.5-VL-7B-SFT, obtained by directly fine-tuning the same backbone on the same training set. The former establishes the pretrained starting point, while the latter provides a data-matched control for isolating gains attributable to \methodname{} rather than exposure to the training data. Vision-R1 \citep{huang2026visionr1} represents text-based visual reasoning, while PixelReasoner \citep{wang2025pixelreasoner} interleaves textual reasoning with image operations to actively acquire additional visual evidence during inference. We further compare with CoVT, LVR, and LIVR \citep{qin2026covt,li2026lvr,li2026livr} as representative latent visual reasoning methods. For local evaluation, we use the official publicly released checkpoints of Vision-R1, PixelReasoner, CoVT, and LVR. We train LIVR on the same training set as \methodname{}. Detailed training configurations for our default \methodname{} model, Qwen2.5-VL-7B-SFT, and LIVR are provided in Appendix~\ref{app:training_configurations}.

\noindent\textbf{Evaluation configurations.}
We evaluate CoVT, LVR, and LIVR with different latent budgets of 8 and 16. For LVR, we can simply set the number of reasoning steps directly. For CoVT, we use two publicly released checkpoints: CoVT-Seg with 8 steps and CoVT-Seg-Depth-DINO with 16 steps. Because LIVR fixes its latent budget during training, we train separate models with 8 and 16 latent tokens on the same training set. Our main \methodname{} configuration uses a total latent budget of $E+S=8$. We evaluate all open models with our evaluation scripts, which follow the official evaluation configurations of the respective benchmarks. For a fair comparison, every evaluated baseline and \methodname{} use the same scripts and benchmark-specific configurations. Detailed evaluation configurations are provided in Appendix~\ref{app:evaluation_configurations}. For the proprietary models, results on CVBench, BLINK, RWQA, and V* are taken from CoVT \citep{qin2026covt}; their POPE scores follow \citet{sarkar2025dpa} for GPT-4o and the Moondream benchmark \citep{m87labs2026moondream} for Claude-4-Sonnet. Since CoVT does not report CVBench results separately for its 2D and 3D subsets, we use its reported GPT-4o and Claude-4-Sonnet results to calculate separate 2D and 3D scores for each model. The calculation is provided in Appendix~\ref{app:cvbench_score_calculation}.

\begin{table}[t]
  \caption{Overall performance across five visual reasoning benchmarks.}
  \label{tab:main_results}
  \centering
  \setlength{\tabcolsep}{2.5pt}
  \begin{tabular*}{\linewidth}{@{\extracolsep{\fill}}lrrrrrrr@{}}
    \toprule
    & \multicolumn{2}{c}{CVBench} & & & & & \\
    \cmidrule(lr){2-3}
    Method & 2D & 3D & BLINK & RWQA & V* & POPE & Avg. \\
    \midrule
    \multicolumn{8}{@{}l}{\textit{Proprietary models}} \\
    GPT-4o$^{\dagger}$ & 74.6 & 83.9 & 63.0 & 69.7 & 42.9 & 85.6 & 70.0 \\
    Claude-4-Sonnet$^{\dagger}$ & 73.5 & 79.1 & 39.6 & 63.7 & 15.2 & 84.6 & 59.3 \\
    \midrule
    \multicolumn{8}{@{}l}{\textit{Base and data-matched models}} \\
    Qwen2.5-VL-7B & 75.4 & 73.3 & 55.6 & 68.9 & 76.6 & 85.0 & 72.5 \\
    Qwen2.5-VL-7B-SFT & 74.2 & 76.9 & 56.3 & 69.5 & 77.5 & 87.6 & 73.7 \\
    \midrule
    \multicolumn{8}{@{}l}{\textit{Explicit/interleaved reasoning}} \\
    Vision-R1 & 74.2 & 77.2 & 52.6 & 68.8 & 78.5 & 88.6 & 73.3 \\
    PixelReasoner & \underline{75.8} & 78.6 & 55.5 & 67.1 & 76.4 & 86.8 & 73.4 \\
    \midrule
    \multicolumn{8}{@{}l}{\textit{Latent visual reasoning}} \\
    LVR (steps=8) & 75.3 & 81.9 & 55.6 & 64.7 & 78.4 & 84.3 & 73.4 \\
    LVR (steps=16) & 74.8 & 81.4 & 52.8 & 62.5 & \underline{80.1} & 80.1 & 72.0 \\
    CoVT (steps=8) & 74.5 & 81.6 & 56.3 & 68.4 & 77.5 & 88.6 & 74.5 \\
    CoVT (steps=16) & 74.3 & \underline{83.9} & 56.8 & \underline{70.1} & 78.4 & 88.5 & 75.3 \\
    LIVR (steps=8) & 73.8 & 83.2 & 56.7 & 69.1 & 79.4 & 87.7 & 75.0 \\
    LIVR (steps=16) & 74.6 & 83.7 & \underline{57.4} & 69.7 & 78.6 & \underline{88.9} & \underline{75.5} \\
    \midrule
    \multicolumn{8}{@{}l}{\textit{Ours}} \\
    \textbf{\methodname{}} & \textbf{78.2} & \textbf{87.8} & \textbf{63.0} & \textbf{71.5} & \textbf{80.6} & \textbf{90.2} & \textbf{78.6} \\
    \textit{$\Delta$ vs. best competitor} & $+2.4$ & $+3.9$ & $+5.6$ & $+1.4$ & $+0.5$ & $+1.3$ & $+3.1$ \\
    \bottomrule
  \end{tabular*}
  \par\vspace{2pt}
  \parbox{\linewidth}{Bold/underline: best/second best among open models. $\Delta$ denotes the absolute difference from the second-best open model in each column. Proprietary-model results were obtained under different evaluation configurations and are included for reference only. $^{\dagger}$: externally reported.}
\end{table}

\noindent\textbf{Results.}
Table~\ref{tab:main_results} shows that \methodname{} achieves the highest average score of 78.6 and the best result on all six reported metrics among the evaluated models. Compared with data-matched SFT, it improves every metric and raises the average by 4.9, showing that the gain cannot be explained solely by exposure to CoVT-Dataset. At the same latent budget of 8, \methodname{} outperforms the strongest latent visual reasoning baseline, LIVR, by 3.6 on average. Simply doubling the latent budget does not yield consistent gains: LVR degrades, while CoVT and LIVR improve only modestly. Even the strongest 16-step result remains 3.1 below \methodname{} with 8 latents. These results suggest that organizing a fixed latent budget around information complementarity is more effective than increasing the latent count alone. The largest improvements occur on CVBench-3D and BLINK, where predictions depend heavily on spatial structure and fine-grained perceptual cues. Improvements on V* and POPE further suggest that the benefits extend to visual search and reliable object grounding.

\subsection{Quantitative Analysis}
\label{sec:quantitative_analysis}

\begin{figure}[t]
  \centering
  \includegraphics[width=\linewidth]{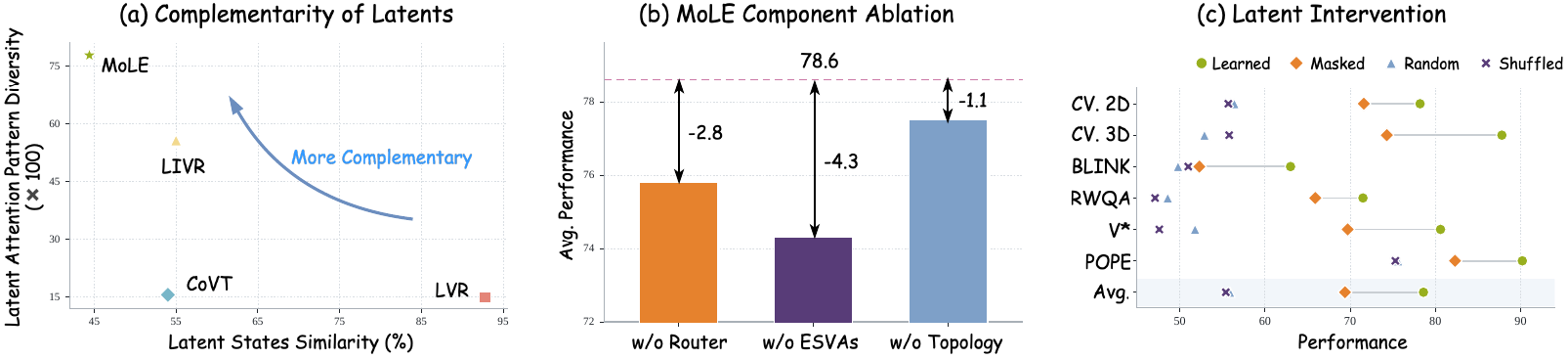}
  \caption{Complementarity, component ablations, and latent-state interventions. (a) Lower latent-state similarity and higher attention-pattern diversity indicate more complementary latent representations. (b) Performance after independently removing each component; the dashed line marks the full \methodname{} result. (c) Performance under inference-time interventions on the learned latent states.}
  \label{fig:complementarity_component_intervention}
\end{figure}

\noindent\textbf{Where do the performance gains come from?}
Starting from the default configuration, we independently retrain three variants that remove sparse visual routing, ESVAs, or topological latent aggregation (Figure~\ref{fig:complementarity_component_intervention}(b)). Without the router, every visual token is made equally accessible to all visual experts. Removing ESVAs retains the shared value projection but eliminates the subsequent expert-specific transformations. Removing topological latent aggregation instead restores conventional causal attention among visual experts, allowing their representations to mix directly. ESVA produces the largest performance drop when removed, indicating that expert-specific visual transformations and the resulting complementary latent representations are particularly important for visual reasoning. The router also contributes substantially by assigning different visual evidence to different visual experts and strengthening their complementarity. Finally, the degradation under causal expert mixing indicates that premature interaction can blur complementary information before it reaches the summary experts, ultimately reducing model performance.

\noindent\textbf{Are the latents used for answer generation?}
We intervene on the learned latent states at inference in three ways: replacing them with random values, shuffling the values within each latent vector, or masking the latent key--value states from answer tokens (Figure~\ref{fig:complementarity_component_intervention}(c)). All three interventions substantially degrade performance. Randomization and shuffling cause the largest drops, but they also introduce strong distribution shifts: meaningless latent states may actively disrupt the decoder's normal language modeling. Masking provides a cleaner test. Even though answer tokens retain direct access to the visual tokens, removing access to the latent pathway lowers the average by 9.2, showing that the learned latents carry information that is used for answer generation.

\noindent\textbf{Do the latents become more complementary?}
We assess latent redundancy from two perspectives. First, pairwise squared cosine similarity between latent-token hidden states measures their numerical similarity. Second, pairwise KL diversity between the visual-token attention patterns of latents measures how differently they attend to visual information. Both metrics are aggregated over the evaluation data, scaled by 100. As shown in Figure~\ref{fig:complementarity_component_intervention}(a), \methodname{} achieves the lowest latent similarity and the highest attention-pattern diversity among the compared latent visual reasoning methods. Detailed formulations of both metrics are provided in Appendix~\ref{app:latent_representation_metrics}.

\subsection{Ablation Studies}
\label{sec:ablation_studies}

\begin{table}[t]
  \caption{Latent-allocation, training-stage, and training-objective ablations.
  (a) Allocation of visual ($E$) and summary ($S$) experts. (b) Training
  stages and objectives.}
  \label{tab:compact_ablations}
  \centering
  \begin{minipage}[t]{0.43\linewidth}
    \centering
    \textbf{(a) Latent allocation}\par
    \vspace{3pt}

    \setlength{\tabcolsep}{3pt}
    \begin{tabular*}{\linewidth}{@{\extracolsep{\fill}}crrr@{}}
      \toprule
      Budget & $E$ & $S$ & Avg. \\
      \midrule
      \multirow{4}{*}{8}
        & 4 & 4 & \textbf{78.6} \\
        & 6 & 2 & 77.2 \\
        & 7 & 1 & \underline{77.4} \\
        & 8 & 0 & 75.7 \\
      \midrule
      \multirow{5}{*}{16}
        & 8  & 8 & \textbf{77.8} \\
        & 12 & 4 & 76.8 \\
        & 14 & 2 & 77.0 \\
        & 15 & 1 & \underline{77.4} \\
        & 16 & 0 & 75.4 \\
      \bottomrule
    \end{tabular*}
  \end{minipage}
  \hfill
  \begin{minipage}[t]{0.53\linewidth}
    \centering
    \textbf{(b) Stages and objectives}\par
    \vspace{3pt}

    \setlength{\tabcolsep}{3pt}
    \renewcommand{\arraystretch}{1.25}
    \begin{tabular*}{\linewidth}{@{\extracolsep{\fill}}llr@{}}
      \toprule
      Ablation & Configuration & Avg. \\
      \midrule
      \multirow{3}{*}{\shortstack{Training\\stage}}
        & Stage~1 only & 69.0 \\
        & Stage~2 only & \underline{75.7} \\
        & Stage~1 + Stage~2 & \textbf{78.6} \\
      \midrule
      \multirow{4}{*}{\shortstack{Training\\objective}}
        & CE only & 76.0 \\
        & CE + MoE & 76.9 \\
        & CE + Diversity & \underline{77.1} \\
        & CE + MoE + Diversity & \textbf{78.6} \\
      \bottomrule
    \end{tabular*}
  \end{minipage}\par
  \vspace{2pt}

  \parbox{\linewidth}{}
\end{table}

\noindent\textbf{Latent allocation.}
We vary the allocation of visual and summary experts under fixed latent budgets of 8 and 16 (Table~\ref{tab:compact_ablations}(a)). The balanced configurations, $E=4,S=4$ and $E=8,S=8$, perform best within their respective budgets. However, the best configuration with 16 latents does not outperform its 8-latent counterpart, indicating that simply increasing the latent budget does not necessarily improve visual reasoning performance. We also set $S=0$ at both budgets to test whether summary experts are necessary. Although these variants devote the entire latent budget to visual experts, their average performance drops by 2.9 and 2.4, respectively. This result shows that allocating capacity to summary experts, which integrate the complementary representations produced by visual experts, is needed.

\noindent\textbf{Training stages and objectives.}
Table~\ref{tab:compact_ablations}(b) examines both the training pipeline and its auxiliary objectives. All stage variants retain the other settings of the default \methodname{} model and are trained for three epochs in total: Stage~1 only uses Stage~1 for all three epochs, Stage~2 only uses Stage~2 for all three epochs, and Stage~1 + Stage~2 follows the default schedule of two Stage~1 epochs and one Stage~2 epoch. The complete two-stage pipeline outperforms either stage alone. Stage~1 shows that the latent pathway can preserve substantial visual information under the visual attention bottleneck, while the gain over Stage~2-only training supports first making this pathway informative before restoring direct visual access. Regarding the objectives, the MoE loss improves expert utilization and router stability, while the diversity loss encourages distinct ESVA transformations; combining both yields the best performance.

\noindent\textbf{More ablations.}
More ablations and results across random seeds are provided in Appendix~\ref{app:more_ablations}.

\section{Conclusion}
\label{sec:conclusion}

We introduced \methodname{}, a framework that organizes latent visual reasoning around complementary latent computation rather than latent count. \methodname{} routes visual evidence to isolated latent visual experts, applies expert-specific visual transformations to routed visual evidence through ESVAs, and aggregates the resulting complementary latent representations with latent summary experts. Across five benchmarks, \methodname{} improves data-matched SFT by 4.9 points on average and outperforms the evaluated latent reasoning methods. Ablations validate the contributions of the architecture and training objectives. Representation analyses show lower latent-state similarity and more diverse visual-attention patterns of latents than other latent visual reasoning methods. Inference-time interventions further confirm the usefulness of our latents.

\subsection*{AI use statement}
In our work, AI-assisted tools were only used for minor language editing and polishing. We reviewed the resulting text and take full responsibility for all the content of the paper.

\subsection*{Reproducibility statement}
We train \methodname{} from the open-source Qwen2.5-VL-7B checkpoint \citep{bai2025qwen25vl} using a 63K-example subset constructed from the open-source CoVT-Dataset \citep{qin2026covt}. To ensure full reproducibility, we provide the complete anonymous code base in supplementary material.

% ...

\bibliography{references}
\bibliographystyle{iclr2027_conference}

\appendix
\section{Appendix}
\label{app:appendix}

\subsection{Method Details}
\label{app:method_details}

This section provides further implementation details for the ESVA and the two-stage attention bottleneck.

\subsubsection{Expert-Specific Value Adapter}
\label{app:esva_details}
ESVA applies a layer-specific normalization followed by an expert-specific two-layer MLP to the output of the shared value projection. At decoder layer $\ell$, the hidden state of visual token $v$ is first processed by the backbone's standard value projection:
\begin{equation}
\mathbf{u}_{\ell,v}
=\operatorname{flatten}\!\left(
\mathbf{W}^{V}_{\ell}\mathbf{h}^{(\ell-1)}_v
\right)
\in\mathbb{R}^{d_V},
\qquad d_V=H_{\mathrm{KV}}d_h,
\label{eq:appendix_shared_value}
\end{equation}
where $H_{\mathrm{KV}}$ and $d_h$ denote the number and dimension of the key--value heads, respectively. This shared projection is retained for every expert. We normalize its output using a layer-specific RMSNorm \citep{zhang2019rmsnorm}:
\begin{equation}
\widehat{\mathbf{u}}_{\ell,v}
=\boldsymbol\omega_\ell\odot
\frac{\mathbf{u}_{\ell,v}}
{\sqrt{d_V^{-1}\lVert\mathbf{u}_{\ell,v}\rVert_2^2
+\varepsilon_{\mathrm{norm}}}},
\label{eq:appendix_esva_norm}
\end{equation}
where $\boldsymbol\omega_\ell\in\mathbb{R}^{d_V}$ is learned independently for each decoder layer and $\varepsilon_{\mathrm{norm}}$ ensures numerical stability.

Each latent visual expert $e$ then applies a bias-free two-layer MLP with hidden dimension $d_b$:
\begin{equation}
\boldsymbol\delta_{\ell,v,e}
=\mathbf{U}_e\,
\operatorname{SiLU}\!\left(
\mathbf{D}_e\widehat{\mathbf{u}}_{\ell,v}
\right),
\qquad
\mathbf{D}_e\in\mathbb{R}^{d_b\times d_V},
\quad
\mathbf{U}_e\in\mathbb{R}^{d_V\times d_b}.
\label{eq:appendix_esva_mlp}
\end{equation}
Here, SiLU denotes the sigmoid-weighted linear unit activation \citep{elfwing2018silu}. The matrices $\mathbf{D}_e$ and $\mathbf{U}_e$ are specific to expert $e$ but shared across decoder layers. In the default configuration, $d_b=32$. For an expert--visual-token connection selected by the router, the MLP output is combined with the original projected value using a learned layer- and expert-specific scale $\gamma_{\ell,e}$:
\begin{equation}
\widetilde{\mathbf{u}}_{\ell,v,e}
=\boldsymbol\delta_{\ell,v,e}
+\gamma_{\ell,e}\mathbf{u}_{\ell,v},
\qquad e\in\mathcal{S}_{\ell,v}.
\label{eq:appendix_esva_output}
\end{equation}
The resulting $\widetilde{\mathbf{u}}_{\ell,v,e}$ replaces the standard value on that routed connection during attention-weighted value aggregation. Thus, ESVA does not replace the backbone value projection; it performs an additional expert-specific transformation of its output. The key projection remains unchanged, and all other attention interactions use the standard value representation.

\subsubsection{Two-Stage Attention Bottleneck}
\label{app:attention_bottleneck_details}
\begin{figure}[t]
  \centering
  \includegraphics[width=\linewidth]{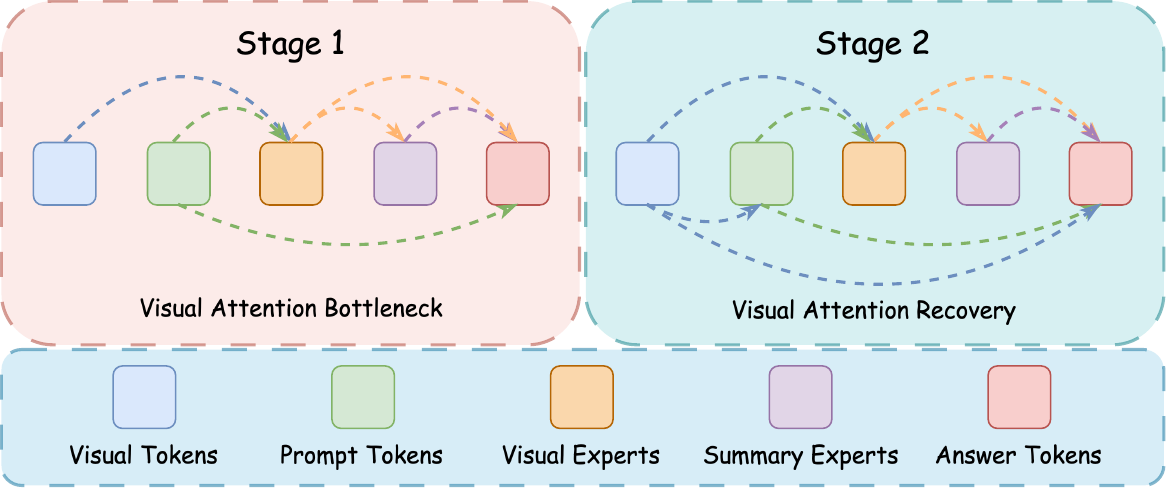}
  \caption{Attention pathways in the two-stage training pipeline. Dashed arrows denote permitted information flow between token groups.}
  \label{fig:attention_bottleneck}
\end{figure}

Figure~\ref{fig:attention_bottleneck} shows the complete information pathways used in the two training stages. In both stages, latent visual experts access visual tokens through router-selected connections and remain isolated from one another, while latent summary experts aggregate only the visual experts. In Stage~1, prompt and answer tokens cannot directly access visual tokens, so visual evidence must reach the answer through the latent pathway. This bottleneck encourages the latent experts to carry information useful for answer generation. Stage~2 starts from the Stage~1 checkpoint and restores direct visual access for prompt and answer tokens, allowing the answer to combine the learned latent representations with fine-grained visual evidence. The two stages optimize the same parameters with the same objective, and all MoLE-specific pathways remain unchanged.

\subsection{Dataset Construction}
\label{app:dataset_construction}

We construct the training set from CoVT-Dataset \citep{qin2026covt}. The following sections describe the source data, our task-balanced sampling strategy, and the online decontamination procedure used to produce the final 63K examples.

\subsubsection{Source Dataset}
CoVT-Dataset aggregates vision-centric subsets of LLaVA-OneVision \citep{li2025llavaonevision} with a re-filtered TallyQA subset \citep{acharya2019tallyqa} and relative-depth questions derived from ADE20K \citep{zhou2017ade20k}. It covers a broad range of tasks, including counting, compositional and spatial reasoning, visual mathematics, OCR, knowledge-based VQA, and detailed image description.

The CoVT-Dataset snapshot used in our experiments contains 889,489 records. Because a record may contain multiple rounds, these records comprise 1,693,665 question--answer pairs in total. The distribution is highly imbalanced: the TallyQA source accounts for more than 40\% of the records, whereas several spatial, depth, OCR, and safety-related sources are substantially smaller. Uniform sampling would largely preserve this skew, so we instead construct a task-balanced subset.

\subsubsection{Task-Balanced Sampling}
We organize the source records into ten high-level task families and allocate 6,300 examples to each family. Table~\ref{tab:dataset_composition} summarizes the resulting composition. For task families supported by multiple sources, we balance those sources where their capacities permit. When a source does not contain enough examples to meet an equal quota, we retain all available examples without replacement and fill the remaining quota from another source within the same task family. For every selected record, we use only the first question--answer pair as the training example.

\begin{table}[t]
  \caption{Composition of the task-balanced 63K training set. Counts separated
  by ``+'' correspond to the listed sources in the same order.}
  \label{tab:dataset_composition}
  \centering
  \setlength{\tabcolsep}{2pt}
  \begin{tabular}{@{}p{0.25\linewidth}p{0.54\linewidth}r@{}}
    \toprule
    Task family & Source group & Samples \\
    \midrule
    Visual math reasoning
      & MathV360K identity group \citep{shi2024mathllava}; IconQA \citep{lu2021iconqa} & $3{,}150+3{,}150$ \\
    Inferential VQA
      & ALLaVA-LAION; ALLaVA-VFLAN \citep{chen2024allava} & $3{,}150+3{,}150$ \\
    Knowledge VQA
      & A-OKVQA \citep{schwenk2022aokvqa} & $6{,}300$ \\
    Compositional reasoning
      & CLEVR \citep{johnson2017clevr} & $6{,}300$ \\
    Safety classification
      & Hateful Memes \citep{kiela2020hatefulmemes} & $6{,}300$ \\
    Detailed captioning
      & Image textualization; numeric-ID group; \texttt{sa}; ShareGPT-4o \citep{cui2024sharegpt4o}
      & $4\!\times\!1{,}575$ \\
    Scene-text OCR
      & ST-VQA \citep{biten2019stvqa} & $6{,}300$ \\
    Counting
      & TallyQA \citep{acharya2019tallyqa} & $6{,}300$ \\
    Multiple-choice VQA
      & Visual7W \citep{zhu2016visual7w} & $6{,}300$ \\
    Spatial understanding
      & VSR \citep{liu2023vsr}; ADE20K relative depth \citep{zhou2017ade20k} & $2{,}152+4{,}148$ \\
    \midrule
    Total & & $63{,}000$ \\
    \bottomrule
  \end{tabular}
\end{table}

After assigning equal quotas to the ten task families, we further divide each family by question format or answer category and distribute its samples across these subtypes. Specifically, we stratify visual mathematics by direct, fill-in-the-blank, and multiple-choice questions; CLEVR by answer type; A-OKVQA by whether the answer contains a rationale; ST-VQA by numeric versus non-numeric answers; TallyQA by count range; and binary or multiple-choice tasks by their class or option labels. When a subtype lacks enough examples, we retain all of its available samples without replacement and fill the remaining quota with other subtypes from the same family. These labels are used only for sampling; the original answer text is preserved as the supervision target.

\subsubsection{Online Sample Decontamination}
We perform decontamination online as part of the task-balanced sampling procedure described above. Whenever the sampler draws a record, we immediately compare it against the samples in POPE, BLINK, V*Bench, RealWorldQA, and CV-Bench and decide whether to retain or discard it. A discarded record does not count toward its task-family quota; instead, we draw a replacement using the same sampling rule. Invalid records and unreadable images are likewise skipped and replaced.

The matching pipeline proceeds from exact matching to candidate retrieval and pairwise verification. First, we compare the file SHA-256 and normalized-pixel SHA-256 of the sampled image with those of all benchmark images. A match from either hash causes the record to be discarded immediately. Records with no exact match proceed to candidate retrieval.

We narrow the subsequent comparison set using three retrieval signals. The first two are pHash and dHash, with maximum Hamming distances of 12 and 8, respectively. The third uses DINOv2 features \citep{oquab2024dinov2} from six image views: one full-image view, one 80\% center crop, and four 70\% corner crops. We retrieve the top 20 benchmark images for each view and take their union. If the normalized question text exactly matches a benchmark question, we also include its associated images and expand DINOv2 retrieval to the top 100. These signals only select benchmark candidates for further verification and do not directly cause a record to be discarded.

We first apply whole-image verification to each retrieved candidate. The images are resized to $256\times256$, and a match requires all three conditions: an aspect-ratio difference of at most 0.02, a color mean absolute error (MAE) of at most 0.055, and a grayscale Pearson correlation of at least 0.97. If all three conditions hold for any candidate, the sampled record is discarded.

Candidates that do not satisfy the whole-image conditions undergo local-feature and geometric verification. We extract up to 2,500 SIFT features and apply the Lowe ratio test \citep{lowe2004distinctive} after limiting the longest image side to 768 pixels. We use a contrast threshold of 0.02 and retain matches with a ratio below 0.75. RANSAC \citep{fischler1981random} estimates the geometric alignment with a 4-pixel reprojection threshold, at most 2,000 iterations, and confidence 0.995. A geometric match requires at least 10 inliers, an inlier ratio of at least 0.28, coverage of at least 0.025 in both images and at least 0.12 in one image, a median reprojection error of at most 3.5 pixels, and an aligned overlap of at least 0.035. For aligned-pixel verification, we erode the valid mask with a $5\times5$ kernel and require at least 256 valid pixels. The aligned region must further satisfy either a grayscale correlation of at least 0.90 with an MAE of at most 0.12, a correlation of at least 0.97, or an MAE of at most 0.055. If any retrieved candidate satisfies the complete geometric and aligned-pixel conditions, the sampled record is discarded; otherwise, it is retained in the training set.

\subsection{Training Configurations}
\label{app:training_configurations}

We describe the training configurations of the default \methodname{} model and the two data-matched baselines, Qwen2.5-VL-7B-SFT and LIVR. All three models use Qwen2.5-VL-7B as the backbone and are trained on the same 63K-example dataset. We first present their shared settings and then specify the differences in trainable parameters, objectives, and training schedules.

\subsubsection{Shared Training Setup}
For all three models, we keep the vision encoder and visual projector frozen and adapt the language backbone with LoRA. LoRA is applied to the attention projections \texttt{q\_proj}, \texttt{k\_proj}, \texttt{v\_proj}, and \texttt{o\_proj}, and to the MLP projections \texttt{gate\_proj}, \texttt{up\_proj}, and \texttt{down\_proj}. We use rank 16, scaling factor 32, and dropout 0.05.

Each image is processed independently with a visual-token budget between 64 and 8192, corresponding to a pixel budget of $[64\times28\times28,\,8192\times28\times28]$. For multi-image examples, this budget is applied separately to every image rather than shared across the sample or batch. We use a maximum sequence length of 9,216 tokens. If an answer exceeds the remaining sequence length, we truncate it while retaining the final end-of-sequence token.

The shared optimization settings use AdamW \citep{loshchilov2019adamw} with batch size 8, learning rate $10^{-4}$, $\beta_1=0.9$, $\beta_2=0.999$, $\epsilon=10^{-8}$, weight decay 0.01, and gradient clipping at 1.0. We use BF16 precision, gradient checkpointing, 5\% linear warmup, and cosine learning-rate decay.

\subsubsection{Default \methodname{} Model}
The default configuration uses $E=4$ latent visual experts, $S=4$ latent summary experts, Router Top-$K=2$, and an ESVA hidden dimension of $d_b=32$. The router-prior scale is $\alpha=1.0$, with $\varepsilon=10^{-6}$ for numerical stability, and the Value RMSNorm uses $\varepsilon_{\mathrm{norm}}=10^{-6}$. The latent embeddings, router projections, and ESVA down- and up-projection matrices are initialized from $\mathcal{N}(0,0.02^2)$. The Value RMSNorm weights are initialized to one, and $\gamma_{\ell,e}$ is initialized to 0.01. In addition to the LoRA parameters, we update the embedding rows of the $E+S$ newly introduced latent experts and all newly introduced \methodname{} parameters in both stages.

Stage~1 is trained for two epochs, followed by one epoch of Stage~2. Both stages use the same joint objective, with $\lambda_{\mathrm{balance}}=10^{-3}$, $\lambda_z=10^{-3}$, and $\lambda_{\mathrm{div}}=10^{-2}$. The complete training pipeline takes approximately 20 hours on two PRO 6000D GPUs.

\subsubsection{Qwen2.5-VL-7B-SFT}
The data-matched SFT baseline uses the same LoRA configuration as the default \methodname{} model and introduces no latent parameters. It is trained for three epochs on the same dataset using only the answer-token cross-entropy loss. Its visual input processing and optimization settings are otherwise identical to those described above.

\subsubsection{LIVR}
LIVR uses the same backbone and LoRA configuration as the default \methodname{} model, together with a fixed set of learnable latent embeddings. We train separate variants with 8 and 16 latent tokens. Each variant follows the same two-stage schedule as \methodname{}, with two epochs in Stage~1 and one epoch in Stage~2, but is optimized using only the answer-token cross-entropy loss. It uses the same training data, visual input processing, and optimization settings as \methodname{}.

\subsection{Evaluation Configurations}
\label{app:evaluation_configurations}

We use the same benchmark-specific evaluation settings for all locally evaluated models. Table~\ref{tab:evaluation_configs} reports the visual-token budget and metric used for each benchmark. The visual budget is applied independently to every image; in particular, each image in a multi-image BLINK example receives the full per-image budget shown in the table.

All five benchmarks use the same Qwen chat template. The system message is fixed to ``You are a helpful assistant.'' The user message contains the image item(s), in their original order, followed by the benchmark-specific question and response-format instruction.

\begin{table*}[t]
  \caption{Per-image visual-token budgets and reported metrics for the five
  evaluation benchmarks.}
  \label{tab:evaluation_configs}
  \centering
  \setlength{\tabcolsep}{3pt}
  \begin{tabular*}{\textwidth}{@{\extracolsep{\fill}}lcp{0.60\textwidth}@{}}
    \toprule
    Benchmark & Tokens/image & Reported metric \\
    \midrule
    POPE
      & 64--512 & Unweighted mean F1 over the adversarial, popular, and random subsets \\
    BLINK
      & 64--512 & Unweighted mean accuracy over the 14 tasks \\
    V*Bench
      & 256--4,096 & Accuracy over all test examples \\
    RealWorldQA
      & 256--4,096 & Accuracy over all test examples \\
    CV-Bench
      & 256--2,048 & 2D: mean of ADE20K and COCO accuracies; 3D: Omni3D accuracy \\
    \bottomrule
  \end{tabular*}
\end{table*}

\subsection{CVBench Score Calculation}
\label{app:cvbench_score_calculation}

CoVT \citep{qin2026covt} reports four CVBench-related scores for GPT-4o and Claude-4-Sonnet: CVBench, Count, Depth, and Dist. CVBench is the aggregate score over the full benchmark and reflects both 2D and 3D visual spatial understanding. Count corresponds to Object Counting, which asks how many instances of a specified object category appear in an image and evaluates instance recognition, object separation, and counting. Depth corresponds to Depth Order, which asks which of two objects is closer to the camera and evaluates depth perception, front--back ordering, and 3D geometry. Dist. corresponds to Relative Distance, which asks which object is closer to a specified reference object and evaluates relative 3D distance. Table~\ref{tab:covt_cvbench_scores} reproduces the scores reported by CoVT.

\begin{table}[t]
  \caption{CVBench-related scores for GPT-4o and Claude-4-Sonnet as reported by CoVT.}
  \label{tab:covt_cvbench_scores}
  \centering
  \begin{tabular*}{\linewidth}{@{\extracolsep{\fill}}lrrrr@{}}
    \toprule
    Model & CVBench & Count & Depth & Dist. \\
    \midrule
    GPT-4o & 79.2 & 65.6 & 86.7 & 81.0 \\
    Claude-4-Sonnet & 76.3 & 62.2 & 77.7 & 80.5 \\
    \bottomrule
  \end{tabular*}
\end{table}

CV-Bench comprises four tasks: its 2D subset contains Spatial Relationship and Object Counting, whereas its 3D subset contains Depth Order and Relative Distance. We therefore compute the 3D accuracy by averaging the reported Depth and Dist. scores. Let $A_{\mathrm{all}}$, $A_{\mathrm{depth}}$, and $A_{\mathrm{dist}}$ denote the reported CVBench, Depth, and Dist. scores, respectively. The 3D accuracy is
\begin{equation}
A_{\mathrm{3D}}
=\frac{A_{\mathrm{depth}}+A_{\mathrm{dist}}}{2}.
\label{eq:cvbench_3d_calculation}
\end{equation}
Because the aggregate CVBench score is the mean of its 2D and 3D scores, we recover the 2D accuracy as
\begin{equation}
A_{\mathrm{2D}}
=2A_{\mathrm{all}}-A_{\mathrm{3D}}.
\label{eq:cvbench_2d_calculation}
\end{equation}
For GPT-4o, these calculations give $A_{\mathrm{3D}}=(86.7+81.0)/2=83.85$ and $A_{\mathrm{2D}}=2\times79.2-83.85=74.55$. For Claude-4-Sonnet, they give $A_{\mathrm{3D}}=(77.7+80.5)/2=79.1$ and $A_{\mathrm{2D}}=2\times76.3-79.1=73.5$. After rounding to one decimal place, we report $(74.6,83.9)$ for GPT-4o and $(73.5,79.1)$ for Claude-4-Sonnet as their respective 2D and 3D results in Table~\ref{tab:main_results}.

\subsection{Latent Representation Metrics}
\label{app:latent_representation_metrics}

We quantify latent complementarity from two perspectives: whether the latent hidden states encode similar representations and whether the latents attend to the same visual evidence. Both metrics are computed per example, averaged over the examples in each benchmark, and then averaged equally across the five benchmarks.

\subsubsection{Latent-State Similarity}
Let $L$ be the number of decoder layers and $M$ the number of latent tokens. For an example $x$, let $\mathbf{h}^{(\ell)}_{x,i}$ denote the hidden state of latent $i$ produced by decoder layer $\ell$, and define its $\ell_2$-normalized representation as
\begin{equation}
\widehat{\mathbf{h}}^{(\ell)}_{x,i}
=\frac{\mathbf{h}^{(\ell)}_{x,i}}
{\lVert\mathbf{h}^{(\ell)}_{x,i}\rVert_2}.
\label{eq:normalized_latent_state}
\end{equation}
We compute the mean pairwise squared normalized inner product among all latent tokens at each layer:
\begin{equation}
s_x^{(\ell)}
=\frac{2}{M(M-1)}
\sum_{1\leq i<j\leq M}
\left(
\widehat{\mathbf{h}}^{(\ell)\top}_{x,i}
\widehat{\mathbf{h}}^{(\ell)}_{x,j}
\right)^2.
\label{eq:latent_similarity_layer}
\end{equation}
The example-level similarity is the mean over all decoder layers:
\begin{equation}
\mathcal{S}_{\mathrm{latent}}(x)
=\frac{1}{L}\sum_{\ell=1}^{L}s_x^{(\ell)}.
\label{eq:latent_similarity_example}
\end{equation}
This score lies in $[0,1]$; a lower value indicates less redundant latent representations.

\subsubsection{Latent Attention-Pattern Diversity}
This metric measures whether different latents distribute their attention over different visual tokens. In a conventional latent visual reasoning method, all $M$ latent tokens can attend to the visual input and are included in this calculation.

Suppose an example contains $V$ visual tokens and each decoder layer has $H$ attention heads. Let $a_{x,\ell,i,h,v}$ be the attention probability assigned by latent $i$ at layer $\ell$ and head $h$ to visual token $v$. We first average over attention heads:
\begin{equation}
\bar{a}_{x,\ell,i,v}
=\frac{1}{H}\sum_{h=1}^{H}a_{x,\ell,i,h,v}.
\label{eq:latent_attention_head_average}
\end{equation}
We then accumulate the resulting visual attention across decoder layers:
\begin{equation}
A_{x,i,v}
=\sum_{\ell=1}^{L}\bar{a}_{x,\ell,i,v}.
\label{eq:latent_attention_layer_aggregation}
\end{equation}
We add $\varepsilon=10^{-8}$ for numerical stability and normalize over the visual-token positions to obtain an attention distribution for each latent:
\begin{equation}
p_{x,i,v}
=
\frac{A_{x,i,v}+\varepsilon}
{\sum_{u=1}^{V}A_{x,i,u}+V\varepsilon}.
\label{eq:latent_attention_distribution}
\end{equation}
For two latents $i$ and $j$, we denote their directed attention-pattern KL divergence by $d_{x,i\rightarrow j}$ and compute it using the natural logarithm:
\begin{equation}
d_{x,i\rightarrow j}
=\sum_{v=1}^{V}p_{x,i,v}
\log\!\left(\frac{p_{x,i,v}}{p_{x,j,v}}\right).
\label{eq:latent_attention_kl}
\end{equation}
The example-level diversity is the mean over all ordered pairs of distinct latents, so both directions of each pair are included:
\begin{equation}
\mathcal{D}_{\mathrm{attn}}(x)
=\frac{1}{M(M-1)}
\sum_{\substack{i,j=1\\i\neq j}}^{M}
d_{x,i\rightarrow j}.
\label{eq:latent_attention_diversity}
\end{equation}
A larger value indicates that the latents attend to visual evidence in more distinct patterns. Finally, for either example-level metric $m$, let $\mathcal{X}_b$ denote the examples from benchmark $b$ and $\mathcal{B}$ the set of five benchmarks. We aggregate and report the metric as
\begin{equation}
\bar{m}_b
=\frac{1}{|\mathcal{X}_b|}\sum_{x\in\mathcal{X}_b}m(x).
\label{eq:latent_metric_benchmark_average}
\end{equation}
We then average the benchmark-level values and apply the reporting scale:
\begin{equation}
m_{\mathrm{report}}
=\frac{100}{|\mathcal{B}|}\sum_{b\in\mathcal{B}}\bar{m}_b.
\label{eq:latent_metric_reported_value}
\end{equation}
Thus, each benchmark contributes equally to the final value, and the plotted scale is 100 times the corresponding raw metric.

For conventional latent visual reasoning methods, all $M$ latent tokens are used for both metrics. \methodname{} instead contains $E$ latent visual experts and $S$ latent summary experts. We compute latent-state similarity over all $M=E+S$ latent tokens, since both sets form intermediate latent representations. For attention-pattern diversity, we use only the $E$ latent visual experts (i.e., set $M=E$ in the corresponding equations), because the summary experts aggregate the visual experts without directly attending to the original visual tokens.

\subsection{More Ablations}
\label{app:more_ablations}

\subsubsection{Parameter-Matched Shared Adapter}
\label{app:parameter_matched}

ESVAs introduce additional parameters into the model's forward computation. To determine whether the resulting performance gains come merely from increased model capacity, we construct a parameter-matched control based on Qwen2.5-VL-7B. Specifically, we add a shared value adapter after the original value projection in every attention layer and choose its size so that the resulting model has a comparable total parameter count to the default \methodname{} model. Unlike ESVAs, these adapters are not expert-specific and therefore do not specialize the transformation of visual evidence. We fine-tune this model on the same 63K training set used for \methodname{} and evaluate it under the same benchmark configurations.

\begin{table}[t]
  \caption{Comparison with a parameter-matched shared-adapter baseline.}
  \label{tab:parameter_matched}
  \centering
  \setlength{\tabcolsep}{2.5pt}
  \begin{tabular*}{\linewidth}{@{\extracolsep{\fill}}lrrrrrrr@{}}
    \toprule
    & \multicolumn{2}{c}{CVBench} & & & & & \\
    \cmidrule(lr){2-3}
    Method & 2D & 3D & BLINK & RWQA & V* & POPE & Avg. \\
    \midrule
    Qwen2.5-VL-7B
      & \underline{75.4} & 73.3 & 55.6 & \underline{68.9} & \underline{76.6} & 85.0 & 72.5 \\
    Parameter-matched adapter
      & 74.9 & \underline{75.4} & \underline{56.5} & 66.7 & 76.3 & \underline{86.8} & \underline{72.8} \\
    \methodname{}
      & \textbf{78.2} & \textbf{87.8} & \textbf{63.0} & \textbf{71.5} & \textbf{80.6} & \textbf{90.2} & \textbf{78.6} \\
    \bottomrule
  \end{tabular*}
  \par\vspace{2pt}
  \parbox{\linewidth}{POPE reports F1; all other metrics report accuracy. Avg. averages the six metrics. Bold and underlined values are the best and second best, respectively.}
\end{table}

As shown in Table~\ref{tab:parameter_matched}, adding a comparable number of parameters improves the average performance of the backbone by only 0.3, whereas \methodname{} outperforms the parameter-matched model by 5.8 and performs better on every reported metric. This result indicates that the gains of \methodname{} cannot be explained by parameter count alone and instead support the importance of explicitly specializing latent computation.

\subsubsection{Stability Across Random Seeds}
\label{app:random_seed_stability}

To assess whether the gains of \methodname{} are stable with respect to training randomness, we train the default configuration independently with random seeds 42, 45, and 48. All runs use the same training data and model configuration and are evaluated under the same benchmark settings.

\begin{table}[t]
  \caption{Performance of the default \methodname{} model across three random seeds. Mean and sample standard deviation are computed over the three runs.}
  \label{tab:random_seed_stability}
  \centering
  \setlength{\tabcolsep}{1.5pt}
  \begin{tabular*}{\linewidth}{@{\extracolsep{\fill}}lrrrrrrr@{}}
    \toprule
    & \multicolumn{2}{c}{CVBench} & & & & & \\
    \cmidrule(lr){2-3}
    Model & 2D & 3D & BLINK & RWQA & V* & POPE & Avg. \\
    \midrule
    Qwen2.5-VL-7B
      & 75.4 & 73.3 & 55.6 & 68.9 & 76.6 & 85.0 & 72.5 \\
    \midrule
    \methodname{} (seed 42)
      & 78.2 & 87.8 & 63.0 & 71.5 & 80.6 & 90.2 & 78.6 \\
    \methodname{} (seed 45)
      & 77.5 & 87.1 & 61.3 & 72.0 & 79.4 & 89.9 & 77.9 \\
    \methodname{} (seed 48)
      & 78.3 & 86.8 & 62.1 & 71.2 & 79.8 & 89.9 & 78.0 \\
    \midrule
    \textit{Mean $\pm$ std.}
      & $78.0{\pm}0.4$ & $87.2{\pm}0.5$ & $62.1{\pm}0.9$ & $71.6{\pm}0.4$ & $79.9{\pm}0.6$ & $90.0{\pm}0.2$ & $78.2{\pm}0.4$ \\
    \bottomrule
  \end{tabular*}
  \par\vspace{2pt}
  \parbox{\linewidth}{POPE reports F1; all other metrics report accuracy. Avg. averages the six metrics.}
\end{table}

Table~\ref{tab:random_seed_stability} shows that the average performance remains between 77.9 and 78.6 across the three runs, with a mean of 78.2 and a standard deviation of 0.4. Every run outperforms the Qwen2.5-VL-7B backbone on all six metrics, and even the lowest average exceeds the backbone by 5.4. These results indicate that the performance gains of \methodname{} are consistent across the evaluated random seeds.

\subsubsection{Backbone and Scale Extension}
\label{app:backbone_scale_extension}

We further examine whether \methodname{} transfers to a different backbone family and remains effective across model scales. We use Qwen3-VL \citep{bai2025qwen3vl}, whose publicly available checkpoints provide finer-grained model sizes than Qwen2.5-VL, and apply \methodname{} to the 2B, 4B, and 8B variants. For each scale, we independently train three models with random seeds 42, 45, and 48 on the same 63K training set and evaluate them under the same benchmark configurations.

\begin{table}[tbp]
  \caption{Backbone and scale extension on Qwen3-VL. Mean and sample standard deviation are computed over three \methodname{} runs at each model scale.}
  \label{tab:backbone_scale_extension}
  \centering
  \setlength{\tabcolsep}{2pt}

  \textbf{(a) Qwen3-VL-2B}\par
  \vspace{2pt}
  \begin{tabular*}{\linewidth}{@{\extracolsep{\fill}}lrrrrrrr@{}}
    \toprule
    & \multicolumn{2}{c}{CVBench} & & & & & \\
    \cmidrule(lr){2-3}
    Model & 2D & 3D & BLINK & RWQA & V* & POPE & Avg. \\
    \midrule
    Qwen3-VL & 72.8 & 86.6 & 45.7 & 65.6 & 75.4 & 89.3 & 72.6 \\
    \methodname{} (42) & 74.7 & 88.3 & 55.5 & 66.3 & 78.0 & 90.2 & 75.5 \\
    \methodname{} (45) & 73.4 & 87.9 & 55.5 & 66.0 & 77.4 & 89.9 & 75.0 \\
    \methodname{} (48) & 73.9 & 88.4 & 55.7 & 66.7 & 77.8 & 90.0 & 75.4 \\
    \textit{Mean $\pm$ std.} & $74.0{\pm}0.7$ & $88.2{\pm}0.3$ & $55.6{\pm}0.1$ & $66.3{\pm}0.4$ & $77.7{\pm}0.3$ & $90.0{\pm}0.2$ & $75.3{\pm}0.3$ \\
    \bottomrule
  \end{tabular*}
  \par\vspace{5pt}

  \textbf{(b) Qwen3-VL-4B}\par
  \vspace{2pt}
  \begin{tabular*}{\linewidth}{@{\extracolsep{\fill}}lrrrrrrr@{}}
    \toprule
    & \multicolumn{2}{c}{CVBench} & & & & & \\
    \cmidrule(lr){2-3}
    Model & 2D & 3D & BLINK & RWQA & V* & POPE & Avg. \\
    \midrule
    Qwen3-VL & 79.1 & 92.4 & 62.2 & 70.3 & 81.2 & 89.3 & 79.1 \\
    \methodname{} (42) & 79.9 & 93.3 & 63.7 & 71.8 & 84.8 & 90.7 & 80.7 \\
    \methodname{} (45) & 79.6 & 92.6 & 63.2 & 71.2 & 84.1 & 90.1 & 80.1 \\
    \methodname{} (48) & 80.2 & 93.1 & 62.8 & 71.6 & 84.6 & 90.3 & 80.4 \\
    \textit{Mean $\pm$ std.} & $79.9{\pm}0.3$ & $93.0{\pm}0.4$ & $63.2{\pm}0.5$ & $71.5{\pm}0.3$ & $84.5{\pm}0.4$ & $90.4{\pm}0.3$ & $80.4{\pm}0.3$ \\
    \bottomrule
  \end{tabular*}
  \par\vspace{5pt}

  \textbf{(c) Qwen3-VL-8B}\par
  \vspace{2pt}
  \begin{tabular*}{\linewidth}{@{\extracolsep{\fill}}lrrrrrrr@{}}
    \toprule
    & \multicolumn{2}{c}{CVBench} & & & & & \\
    \cmidrule(lr){2-3}
    Model & 2D & 3D & BLINK & RWQA & V* & POPE & Avg. \\
    \midrule
    Qwen3-VL & 80.6 & 93.2 & 64.3 & 69.5 & 82.7 & 88.2 & 79.8 \\
    \methodname{} (42) & 81.3 & 93.7 & 67.1 & 71.8 & 85.3 & 89.9 & 81.5 \\
    \methodname{} (45) & 80.8 & 93.4 & 66.4 & 71.5 & 84.7 & 89.7 & 81.1 \\
    \methodname{} (48) & 80.8 & 93.7 & 66.5 & 72.3 & 84.9 & 90.2 & 81.4 \\
    \textit{Mean $\pm$ std.} & $81.0{\pm}0.3$ & $93.6{\pm}0.2$ & $66.7{\pm}0.4$ & $71.9{\pm}0.4$ & $85.0{\pm}0.3$ & $89.9{\pm}0.3$ & $81.3{\pm}0.2$ \\
    \bottomrule
  \end{tabular*}
  \par\vspace{2pt}
  \parbox{\linewidth}{Parentheses after \methodname{} give the random seed. POPE reports F1; all other metrics report accuracy. Avg. averages the six metrics.}
\end{table}

As shown in Table~\ref{tab:backbone_scale_extension}, \methodname{} raises the mean average score over the corresponding Qwen3-VL backbone by 2.7, 1.3, and 1.5 at the 2B, 4B, and 8B scales, respectively. Every run improves over its corresponding backbone on all six metrics, while the standard deviation of the average score is at most 0.3 at each scale. These results show that the gains remain consistent across random seeds and extend beyond the default Qwen2.5-VL-7B backbone to multiple Qwen3-VL model sizes.

\subsubsection{Routing Density}
\label{app:routing_density_ablation}

Router Top-$K$ controls how many latent visual experts receive each visual token at a decoder layer. Starting from the default \methodname{} configuration, we independently retrain models with $K\in\{1,2,3,4\}$ to examine how the density of expert--visual-token connections affects performance. As shown in Table~\ref{tab:routing_esva_ablation}(a), the default $K=2$ achieves the best average performance. Routing each visual token to only one expert may be overly restrictive, whereas larger values of $K$ increase the overlap in visual evidence received by different experts without providing further gains. This suggests that moderate routing density balances selective evidence allocation with sufficient expert processing.

\subsubsection{ESVA Hidden Dimension}
\label{app:esva_hidden_dimension_ablation}

We vary the hidden dimension $d_b$ of the two-layer ESVA MLP over $\{16,32,64,128\}$ while keeping the remaining default configuration unchanged. This ablation tests whether the benefit of ESVA depends on increasing its transformation capacity. Table~\ref{tab:routing_esva_ablation}(b) shows that $d_b=32$ performs best, while larger hidden dimensions do not improve performance. Thus, the ESVA gains do not arise from simply increasing adapter capacity, and a compact hidden dimension is sufficient for expert-specific visual transformation.

\begin{table}[tbp]
  \caption{Ablations of (a) routing density and (b) ESVA hidden dimension. Only average performance is reported.}
  \label{tab:routing_esva_ablation}
  \centering
  \begin{minipage}[t]{0.46\linewidth}
    \centering
    \textbf{(a) Routing density}\par
    \vspace{3pt}

    \setlength{\tabcolsep}{4pt}
    \renewcommand{\arraystretch}{1.1}
    \begin{tabular*}{\linewidth}{@{\extracolsep{\fill}}lr@{}}
      \toprule
      Router Top-$K$ & Avg. \\
      \midrule
      $K=1$ & 77.8 \\
      $K=2$ & \textbf{78.6} \\
      $K=3$ & 77.4 \\
      $K=4$ & \underline{77.9} \\
      \bottomrule
    \end{tabular*}
  \end{minipage}
  \hfill
  \begin{minipage}[t]{0.46\linewidth}
    \centering
    \textbf{(b) ESVA hidden dimension}\par
    \vspace{3pt}

    \setlength{\tabcolsep}{4pt}
    \renewcommand{\arraystretch}{1.1}
    \begin{tabular*}{\linewidth}{@{\extracolsep{\fill}}lr@{}}
      \toprule
      Hidden dimension $d_b$ & Avg. \\
      \midrule
      16  & 77.2 \\
      32  & \textbf{78.6} \\
      64  & 77.0 \\
      128 & \underline{77.4} \\
      \bottomrule
    \end{tabular*}
  \end{minipage}\par
  \vspace{2pt}

  \parbox{\linewidth}{Avg. averages the six metrics. The default uses $K=2$ and $d_b=32$. Bold and underlined values are the best and second best within each ablation, respectively.}
\end{table}

\subsubsection{Inference Cost}
\label{app:inference_cost}

We quantify the inference overhead of \methodname{} relative to its Qwen2.5-VL-7B backbone. For every RWQA example, we evaluate both models and record inference time, floating-point operations (FLOPs), and peak GPU memory, then average each measurement over the full benchmark. Table~\ref{tab:inference_cost} reports the results.

\begin{table}[tbp]
  \caption{Average inference cost on RWQA.}
  \label{tab:inference_cost}
  \centering
  \setlength{\tabcolsep}{6pt}
  \renewcommand{\arraystretch}{1.1}
  \begin{tabular*}{0.92\linewidth}{@{\extracolsep{\fill}}lrrr@{}}
    \toprule
    Model & Time (s) & FLOPs & Peak memory (GiB) \\
    \midrule
    Qwen2.5-VL-7B & 0.374478 & $3.481782{\times}10^{13}$ & 15.836116 \\
    \methodname{}  & 0.460499 & $3.510919{\times}10^{13}$ & 16.081371 \\
    \midrule
    Relative overhead & $+22.97\%$ & $+0.84\%$ & $+1.55\%$ \\
    \bottomrule
  \end{tabular*}
\end{table}

\methodname{} adds only $0.84\%$ FLOPs and $1.55\%$ peak memory, corresponding to $0.245$ GiB. Its measured inference time increases by $22.97\%$, but the absolute increase is $0.086$ seconds per example and the average latency remains below $0.5$ seconds. In return, \methodname{} improves the average benchmark score over the backbone by 6.1 points. We therefore consider this inference overhead acceptable for the resulting performance gain.

\end{document}